\documentclass[preprint, 5p, authoryear,times]{elsarticle}
\pdfoutput=1
\usepackage{lineno,hyperref}
\usepackage{graphics} 
\usepackage{subfig}
\usepackage{mathtools}
\usepackage{microtype}
\usepackage[flushleft]{threeparttable}
\usepackage{algorithm2e}

\modulolinenumbers[5]

\journal{Engineering Applications of Artificial Intelligence}

\usepackage{numcompress}\biboptions{authoryear,semicolon}

\begin{document}

\begin{frontmatter}

\title{An Online Semantic Mapping System for Extending and Enhancing Visual SLAM}

\author{Thorsten Hempel* and Ayoub Al-Hamadi\corref{mycorrespondingauthor}}

\cortext[1]{Corresponding author}
\ead{Thorsten.Hempel@ovgu.de}
\address{Neuro-Information Technology, Otto von Guericke University Magdeburg, 39106 Magdeburg}


\begin{abstract}
We present a real-time semantic mapping approach for mobile vision systems with a 2D to 3D object detection pipeline and rapid data association for generated landmarks. Besides the semantic map enrichment the associated detections are further introduced as semantic constraints into a simultaneous localization and mapping (SLAM) system for pose correction purposes. This way, we are able generate additional meaningful information that allows to achieve higher-level tasks, while simultaneously leveraging the view-invariance of object detections to improve the accuracy and the robustness of the odometry estimation. We propose tracklets of locally associated object observations to handle ambiguous and false predictions and an uncertainty-based greedy association scheme for an accelerated processing time. Our system reaches real-time capabilities with an average iteration duration of 65~ms and is able to improve the pose estimation of a state-of-the-art SLAM by up to 68\% on a public dataset. Additionally, we implemented our approach as a modular ROS package that makes it straightforward for integration in arbitrary graph-based SLAM methods. 
\end{abstract}


\end{frontmatter}

\sloppy
\section{Introduction}
Visual simultaneous localization and mapping (SLAM) takes a key role in many autonomous mobile systems. Traditionally, these methods rely on local geometrical features, such as corners (\citealt{6605544}), lines (\citealt{7989522}) or surface patches (\citealt{rgbd-mapping}), or directly match image intensities (\citealt{engel14eccv}). In either case, the generated map is a pure geometrical representation of the environment. This restricts most SLAM methods to basic navigation and obstacle detection tasks. 
In the past years the idea of extending these maps with additional semantic data emerged with the prospect of enhancing SLAM to a higher level of environment perception and thus, to be capable of engaging in more complex scenarios, \textit{e.g.} in Human-Robot Interaction (HRI)(\citealt{domas}) and object manipulation (\citealt{Rusu2010Semantic3O}).

Recent approaches from \cite{semmap:mccormac},\cite{semmap:nakajima2}, \cite{semmap:nakajmi} and \cite{semmap:sengputa} in this direction combine traditional visual SLAM with CNN-based object detection methods to assign existing map elements with semantic labels.
Other methods, often referred to as 
\textit{object-slam}~(\citealt{slam:nicholson, slam:salas-moreno, slam:yang, slam:mccormac}), entirely rely on semantic object landmarks without the implication of conventional hand-crafted features or direct matching techniques. These \textit{object-based} approaches are challenged with correct object association, accurate object localization and require settings with enough known objects to obtain reliable camera pose information. However, they can reach superior performance compared to traditional approaches in respect of robustness towards errors from large viewpoint changes and surroundings with recurring texture patterns and poor-featured surfaces. Both, traditional and object-based SLAM approaches, require static environments to draw correct conclusions from their photogrammetric methods. Hence, several approaches~(\citealt{dynslam:bescos, dynslam:hempel, pablo, kim}) have been proposed to increase the robustness towards dynamic induced errors to enable the application in non-static environments. Yet, they tend to be very computational costly or require prior knowledge about the scene.  
\begin{figure}[t]
	\centering
	\includegraphics[width=\linewidth]{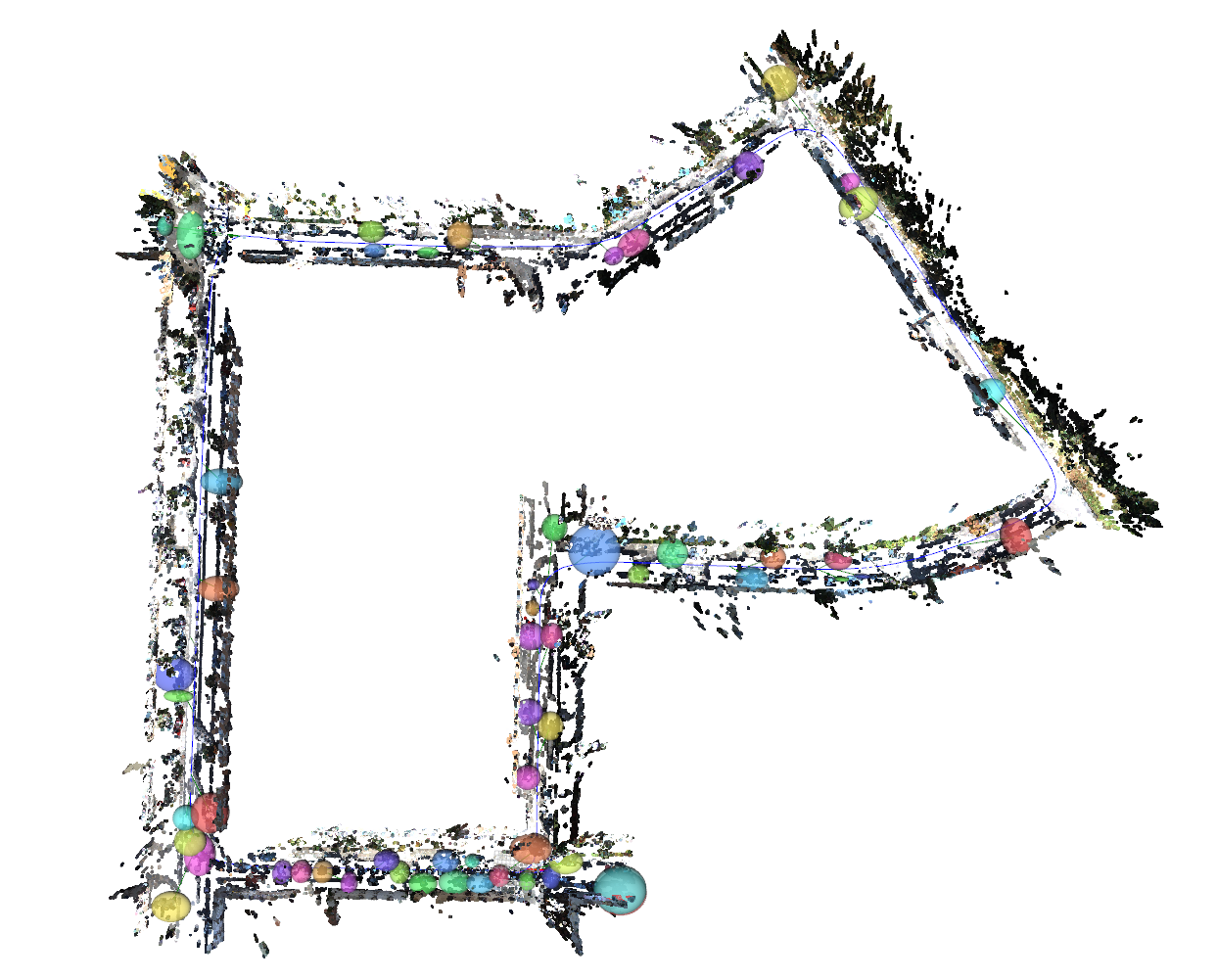}
	
	\caption{Example image of a point cloud map supplemented with 3D object markers that are derived from 2D detections. The markers are mapped as arbitrary colored spheres and bloated in their size by factor three for demonstration purposes.}
	\label{fig:fig1_}
\end{figure}

\sloppy
To tackle the challenge of incorporating semantic meaning into environment maps in a consistent and robust way we propose a hybrid method that aims to combine the advantage of both worlds: the matching precision of traditional geometric features and the reliability of CNN-based semantic landmarks. 
We will show that the enhancing of conventional SLAM methods with semantic object markers expands the visual perception capability for more intelligent tasks and additionally provides utility to significantly reduce errors in the pose estimation. Consequently, we will demonstrate that the integration of semantic objects into an open-source SLAM is mutually benefiting the information value of the map as well as its localization robustness towards dynamic environments and errors in feature recognition. To the best of our knowledge this is the first approach that specifically targets real-time and scaling capabilities to make our proposed method applicable for mobile robotics.
Figure \ref{fig:fig1_} shows an example image point-cloud map generated from a feature-based slam, that is extended with semantic landmarks (shown as arbitrary colored spheres). Our main contributions are:
\begin{itemize}
	\item A tracklet strategy for the efficient and robust selection of meaningful landmark proposals. As dynamic objects can severely harm visual-based localization, we facilitate the generated tracklets to apply a validation step that examines object candidates for dynamic and unsteady behavior.  
	\item A uncertainty-based custom nearest neighbor method for a fast and robust data association process.
	\item The evaluation on the public available TUM RGB-D dataset from~\cite{rgbtum} and the stereo KITTI dataset from~\cite{Geiger2013IJRR} in challenging real world scenarios.
	\item The implementation as an real-time plugin module for a straightforward integration into other online SLAM methods.	
\end{itemize}

The remaining part of the paper is outlined as follows: First, we will review various key approaches of the recent literature that target the inclusion of semantic information into SLAM in section \ref{rel_work}. In section \ref{method} we will give details about our proposed method followed by our a qualitative and quantitative evaluation in section \ref{experiments}. We complete this letter with a conclusion in section \ref{conclusion}.

\section{Related Work}
\label{rel_work}
Using semantic information in SLAM has been studied in a variety of approaches. \cite{semmap:nakajmi, semmap:mccormac, semmap:nakajima2, semmap:nguyen} integrate object information to associate geometric map points with semantic labels. \cite{slam:iqbal, slam:pire} insert complete object instances into the map instead of classifying already obtained map elements. 
These approaches facilitate object recognition methods to integrate semantic information into a geometrical context, but don't make the localization itself benefit from the additional generated semantic data.
A step towards this direction has been presented by \cite{slam:Zhong} who eliminate feature points laying on projected objects of dynamic associated classes. Similarly, \cite{slam:yu} examine and remove feature points located on segmented area that are classified as \textit{people} and therefore assigned as dynamic. Both approaches affect the trajectory estimation only in a passive manner by excluding unreliable data relative to its classification. 
More proactively, \cite{slam:salas-moreno} presented a localization algorithm that is entirely based on object-level landmarks, which are derived from a prebuilt model dataset. Further \textit{object-SLAM} approaches from \cite{slam:yang} and \cite{slam:nicholson} define object landmarks by fitting cuboids and quadrics, respectively, to object bounding boxes provided by neural networks.

More similar to our approach, \cite{slam:bernreiter} address a hybrid solution by fusing semantic submaps with odometry measurements. The submaps are generated via multi hypothesis tracking and associated with an adapted Hungarian algorithm. \cite{slam:li} derive 3D bounding cuboids of objects from a sequence of 2D detections for relocalization from large viewpoint changes. \cite{extsem} extend robotic maps with semantic 3D shape priors that have similar appearance to projected object detections. On the contrary, \cite{slam:bowman} and \cite{Doherty2020ProbabilisticDA} propose to directly incorporate semantic constraints into an probabilistic optimization framework. All of the described methods have in common that they use visual-based neural networks to facilitate semantic information in the SLAM process. These object detectors produce non-Gaussian and discrete random variables and, accordingly, introduce the risk of incorporating false predictions into the data association process. The implication of faulty detections can result increased computational effort or even unrecoverable corrupt the entire pose estimation. While \cite{slam:bowman} inherently neglect false negatives and false positives, \cite{slam:li} only require the same object predictions from both cameras of the stereo setup. \cite{slam:bernreiter} explicitly deal with misdetections as a possible submap option with the drawback of increased computation complexity. Whereas, \cite{slam:nicholson} assume the data association as given. 

In contrast to these approaches we propose a straightforward pre-selection scheme with tracklets of local-associated detections. By suppressing potential semantic landmarks from false object prediction in this upstream process we are able reduce the computation cost and the risk of errors for subsequent processing steps, notably the data association. In addition, we take advantage of the tracklets by introducing a heuristic to detect non-static object behavior. Hence, we reject those object that are potentially harmful for the localization process. 
Unlike \cite{slam:salas-moreno} our system does not require any prior of the objects shape or size. Instead, we iteratively gather and optimize this information along the trajectory. We propose an efficient customized nearest neighbor process to solve the data association where we incorporate the object size as well as an uncertainty factor to reduce the risk of association errors. In return, this gives us the opportunity to utilize our landmark detections to optimize the pose graph and thus, rectify our trajectory in an active manner. 

\section{Method}
\label{method}

\begin{figure*}[]
	\centering
	\includegraphics[width=\linewidth]{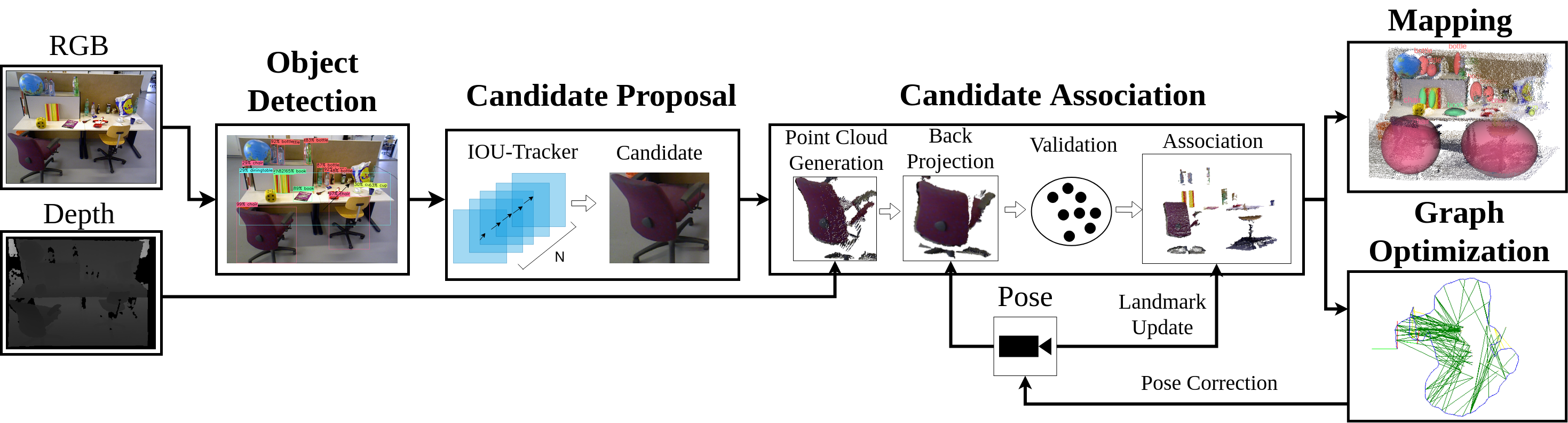}

	\caption{Overview of the proposed method. We predict objects in 2D images and track them using an adapted IOU-Tracker. Confident tracks are proposed as landmark candidates and back projected into world frame. After a validation step each candidate further processed in a data association step. The resulting landmarks are mapped as spheres and facilitated as constraints to improve the pose estimation.}
	\label{method_ga}
\end{figure*}

\newcommand{\subcaption}[1]{
	\BlankLine
	\hspace*{\fill}	#1
		\vspace{.1\baselineskip}
		\hrule	
		\BlankLine
}
\newcommand{\m}[1]{$\boldsymbol{#1}$}
\newcommand{\mm}[1]{${#1}$}
\begin{algorithm}[]
	\SetKwData{Left}{left}\SetKwData{This}{this}\SetKwData{Up}{up}
	\SetKwFunction{Union}{Union}\SetKwFunction{FindCompress}{FindCompress}
	\SetKwInOut{Input}{input}\SetKwInOut{Output}{output}
	\caption{Semantic Landmark Mapping}\label{algorithm}
	\SetAlgoLined
	\Input{Object measurements \m{Z}, tracklets \m{R=\{\}} odometry estimations \m{X}}
	\Output{Landmark map \m{L}}
		\BlankLine
	 \mm{R} $\gets$ generate according to $IOU(Z,R)$ (\ref{jaccard})\;
	 \For(\tcp*[f]{Candidate generation}) {\mm{r} in \mm{R}}{
	 \If{$size(r)$ $=$ \m{min\_trackletsize}}{
		  \mm{C} $\gets$ add new \m{c(r)}\;		  
 }

}

	\For(\tcp*[f]{Association loop}) {\mm{c} in \mm{C_{New}}}{
		Generate point cloud \m{p} for all \mm{z} in \mm{c}\;
		Project \mm{P_{Camera}} to \mm{P_{World}} using \m{X} \;
		Determine cloud centroid points \m{x_p} for all \m{p} in \mm{c}\;
	\BlankLine
		\If(\tcp*[f]{Validate}) {$ \frac{1}{n}\sum \vert x_i - \overline{x} \vert $ $>$ \mm{thresh}}
		{
			\mm{C}$\gets$ remove \mm{c}\;
		}
		\BlankLine
	Estimate 3D centroid \m{X} for \m{c} using (\ref{prob})\;
	 \m{r(c_i)} $\gets$ Validation Gate (\ref{gate}) \;
	 \m{L_R} $\gets$ All \mm{l} in \mm{L} \textbf{where}  \mm{dist(d_l,d_{c_i})}$\leq$\mm{r_{c_i}}\;
	 \If {\mm{L_R} = empty}{
	 	\mm{L} $\gets$ add new landmark \m{l(c_i)}\;
	 }
 	\Else
 	{
 		Nearest Neighbor (\ref{knn}) in \mm{L_R} $\rightarrow$ \mm{l(c_i)}\;
 		\mm{l(c_i)} $\gets$ revise landmark according to \mm{c_i}\;
 	}
	}
\end{algorithm}

In this section our system will be introduced in details as shown in Figure \ref{method_ga}. It can be separated into three different steps: the landmark candidate proposal step, the candidate association step and the object refinement step. In the following, we will describe each processing step. Algorithm \ref{algorithm} shows how these different steps are linked with each other. A single 2D object prediction will be called \textit{measurement}. Multiple associated \textit{measurement} in form a of tracklet proposes a landmark \textit{candidate}, whereas an accepted candidates becomes a \textit{landmark}.  

\subsection{Landmark candidate proposal generation}

Neural networks for object detection~(\citealt{cnn:ren,cnn:redmon, yolo4}) have archived tremendous accuracy and efficiency in recent years. However, their prediction results are highly affected by environmental conditions. Unusual object appearance and insufficient illumination can rapidly decrease the precision rate and increase the number of false positives and misclassification. For mapping purposes, this can lead to the inclusion of not existing or false classified objects, distortions of the data association and finally excessive computational costs. If used for pose estimation the entire trajectory can be irreversible adulterated.  

In order to avoid these degrading effects, we propose to perform a low-level selection process to sort out mispredictions, potential dynamic objects, and other untenable object detections before passing them to the data association step. For this purpose, we follow the track-by-detection paradigm and use an adapted IOU tracker~(\citealt{bochinski}). The tracker associates bounding box
predictions for objects with high Jaccard index $J$ and matching class label over time consecutive frames. In particular, the Jaccard index evaluates the similarity of two predicted bounding boxes $bb_1$ and $bb_2$ by computing the ratio of its overlap and union. 
\begin{equation}
\label{jaccard}
J(bb_1,bb_2)= \frac{bb_1\cap bb_2}{bb_1\cup bb_2}
\end{equation}
Detection pairs from consecutive frames with minor viewpoint changes that have a high Jaccard index are likely to be projected from the same real-world object as their image location and dimensions are similar.
In this way, we incrementally construct local tracklets of spatial and temporal associated object detections where each additional \textit{measurement} increases the confidence in a true positive class label prediction. Furthermore, this procedure helps us to avoid the mapping of dynamic objects as due to its motion of the bounding boxes the Jaccard index is likely to fall below the threshold. If a track length \textit{N} reaches a specific \textit{minimum length} we assume it as a valid detection. At this point, the entire tracklet will be proposed as \textit{landmark candidate} and passed to the next processing step. Detections that are not associable are used to create new tracklets.

\subsection{Landmark candidate association}
\bigbreak
\subsubsection{Candidate Validation}

Having multiple \textit{measurements} of the same objects gives us the opportunity to conduct further analytical procedures. More precisely, we examine each candidate for fidget (\textit{e.g} caused by occlusions) or dynamic behavior. To this end, we first extract a cuboid point cloud for each measurement based on its predicted bounding box size and then back project each point cloud into the world reference. In the following step, we determine the centroid point for each cloud separately and use them to calculate the mean absolute deviation (MAD) of the tracklets measurement centroids. A high MAD indicates position instability and results in the rejection of the candidate. It is commonly introduced by dynamic objects or changing occlusion of the object along the track. In these cases, we don't have enough reliable localization information to create a static landmark.
\subsubsection{Candidate Localization}

For those candidates who pass the check, we search for the centroid of the 3D object represented by $\mathbf{X}=(X,Y,Z)^T$ that is projected onto the image plane at the location $x=f(\mathbf{X})$ with the projection function $f$.
To model its probability we follow the approach described by \cite{10.5555/861369} and \cite{slam:li} and introduce measurement noise as zero-mean Gaussian noise. Accordingly, we receive a measurement model defined as $x=f(\mathbf{X})+\eta$ with $\eta\sim \mathcal{N}=(\mathbf{0},\Sigma)$.\\
The tracklets probability that its measured center points $x_{1:t}=\{x_1, x_2, x_3, ... , x_t\}$ are projected from the 3D object centroid $\mathbf{X}$ can be expressed as follows:

\begin{equation}
p(x_{1:t} \vert \mathbf{X}) =  \frac{exp (-\frac{1}{2}(f(\mathbf{X})-x_{1:t})^T\Sigma^ {-1}(f(\mathbf{X})-x_{1:t})) }{ \sqrt{(2\pi )^{2} \vert \Sigma \vert }}
\end{equation}

The posterior distribution of $\mathbf{X}$ leading to track $x_{1:t}$ according to Bayes is

\begin{equation}
p(\mathbf{X}\vert x_{1:t})=\frac{p(x_{1:t}\vert\mathbf{X})P(\mathbf{X})}{p(x_{1:t})}.
\end{equation}

Assuming a uniform prior distribution and independent measurements we receive

\begin{equation}
p(\mathbf{X}\vert x_{1:t}) = \prod_{t=1}^{T}p(x_t\vert\mathbf{X}),
\end{equation}

where the objective is to find the objects' centroid $\mathbf{X^*}$ that obtains the maximum unnormalized \textit{a posteriori} probability.
\begin{equation}\label{prob}
\mathbf{X^*}=\operatorname*{arg\,max}_\mathbf{X} p(\mathbf{X}\vert x_{1:t})
\end{equation}

To this end, we create point-location hypotheses of our landmark by drawing samples from $p(\mathbf{X}\vert x_{1:t})$ and calculate its probability with \ref{prob}. As sampling technique we first triangulate 3D point from our samples as initial starting point and proceed with a random walk Monte Carlo sampling.

\subsubsection{Landmark Association}
As a result of our preprocessing step for rejecting invalid object predictions our landmark candidates remain with two possible options for data association. The candidate is either a complete new landmark or has been seen and registered as landmark before. To check if the candidate belongs to a known landmark we determine a validation gate within a possible association landmark is expected. The size of the validation gate depends on the certainty about the candidates' location in world reference, which is almost entirely based on the ego-motion estimation. To be independent of the localization algorithm we make the size of the validation gate $r$ relative to the period $\Delta t$, which represents the time in \textit{seconds} since the last association of a known landmark, and the object size $s$ in \textit{meters}.
\begin{equation}
\label{gate}
r = \sqrt{\frac{\Delta t}{u} \cdot s} 
\end{equation}
The longer we couldn't associate a new candidate with a known landmark, the less confident we are with its exact position. Variable $u$ modulates the ground uncertainty on how strong gate size should grow with time. We empirically found that $u=10$ offers the best setting for having the gate size just as big enough for finding our target association landmark. 
Having the validation gate $r$ we can simply calculate the euclidean distance between the candidate and our known landmarks to make a preselection of those landmarks that are laying inside the gate with a matching class label. Ideally, we end up with one possible landmark for association. If we found multiple landmarks from the same class in the candidates' validation gate we have to conduct further processing. 
In this case we claim that the comparison of distances between the centroids to find the best fitting landmark to the candidate is unreliable. We assume that the point cloud representing the object is barely mapping its entire appearance as the camera is only able to perceive the front of the object's surface. Depending on this and the object's size the estimated centroid is only an approximation of the object's real center point. The comparison of the centroids from different viewpoints is therefore insufficient. 
Our approach for solving this problem is to calculate the Euclidean distance between the point clouds from the candidate $c$ and the landmark $l$ within the gate $r$ by using the nearest-neighbor search.
\begin{equation}
\label{knn}
l_t(c)= \arg \min_{l\in L}  \sum_{i=1}^{P_c}d(p^{i}, p_l^{NN}) \cdot \frac{1}{P_c}
\end{equation}
The goal is to find for every point $p$ in the candidate's point cloud $P_c$ its nearest neighbor point $p^{NN}_l$ from the landmark point cloud $P_l$ and, consequently, to find those landmark $l$ from the set of landmarks $L$ that has averagely the shortest distance $d$ between the nearest neighbors.
This brings the advantage to choose rather the landmark with the strongest overlap instead of selecting the shortest distance between the centroids. If we find an appropriate landmark, we associate our candidate with it and reset $\Delta t_r$. In the case of not finding an appropriate association, we assume that the candidate has not been seen before and register it as a new landmark to our map.

\subsection{Pose optimization}
\bigbreak
Seeing the same static object from different viewpoints gives us the ability to make conclusions about our own localization. We exploit this by providing landmark detection information to our SLAM backbone to integrate it as constraints into its pose graph. We assign every landmark an unique id. If we revisit a landmark, we pass its position on the image plane together with the landmark id to the localization algorithm, where its used as additional edge in the pose graph for optimizing the trajectory estimation.

\subsection{Landmark refinement}
\label{revaluation}
\bigbreak
With the optimization of the trajectory by introducing new semantic landmark detection (or other loop closure efforts) we update our landmark positions according to the corrected pose graph. This provides us a way to revise previous association errors. For example, if our estimation drifts apart from the actual trajectory, a new landmark candidate might not find it's association landmark within its validation gate and will be registered as an allegedly new landmark. In this case, we will end up with two landmarks with different IDs from the same real object. After trajectory optimizations and the landmarks will be corrected subsequently and placed on the missed association landmark from before. In this way, we are able to detect excessively overlapped landmarks of the same class and fuse them into one single landmark.
This enables us to map even very large objects by incrementally merging multiple semantic landmarks into one.

\subsection{System architecture}
\begin{figure}[t]
	\centering
	\includegraphics[width=\linewidth]{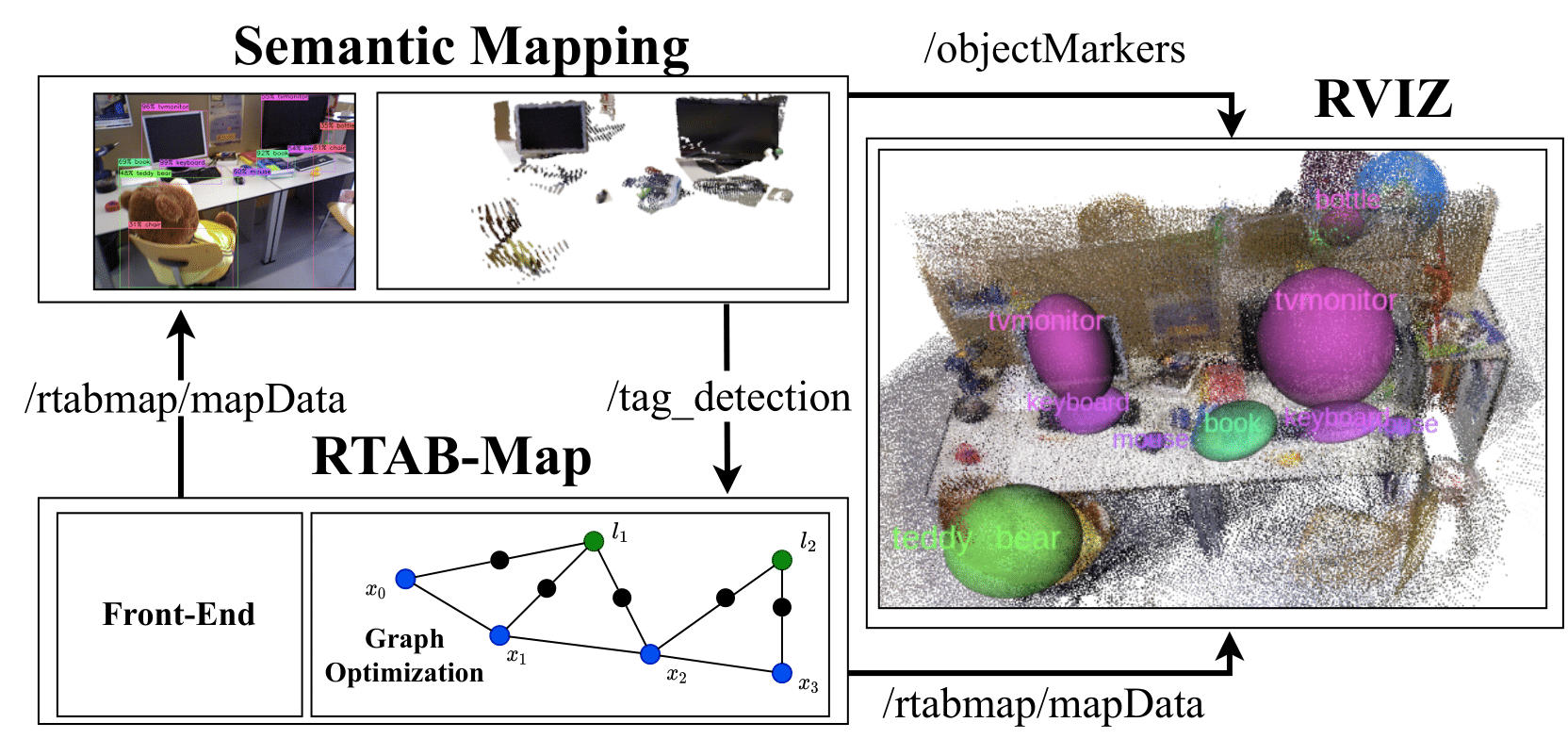}

	\caption{Overview of the implementation of our proposal. Our semantic mapping ROS node communicates with the RTAB-Map SLAM to request pose graph information and provides semantic detections for graph optimization. The generated landmarks are added to the point cloud map in the RVIZ visualization tool.}
	\label{implementation}
\end{figure}

\bigbreak
Our intention for the implementation of our algorithm was to create a flexible enhancement module that can be easily subjoined to other conventional SLAM implementations. Accordingly, we developed our software as a Robot Operating System (ROS)~(\citealt{ros}) package. The communication between this package and the SLAM is administrated by ROS topics, which makes it straightforward to interact with other ROS integrated SLAM implementation. Exemplary, we chose the graph-based RTAB-Map from \cite{slam:rtabmap} as our SLAM backbone as it provides the modularity to easily switch between different types of localization and graph optimization algorithms. An overview of interconnection is provided in Figure \ref{implementation}. We recurrently draw the latest pose graph update from RTAB-Map for projecting and associating or landmarks (see Section \ref{method}) and publish new ascertained semantic landmarks. These landmarks are shaped as AprilTag~(\citealt{apriltag}) messages to be compatible with the standard RTAB-Map. For visualization, we add our semantic landmarks as spherical markers to the point cloud map in RVIZ. The sphere's size is estimated based on the objects point cloud dimensions.

\begin{figure}[b]
	\centering	
	\subfloat[\centering Image from a test sequence with object predictions and corresponding 2D bounding boxes.]
	{\setlength{\fboxrule}{0.5pt}\setlength{\fboxsep}{0pt}\fbox{\includegraphics[width=0.47\linewidth]{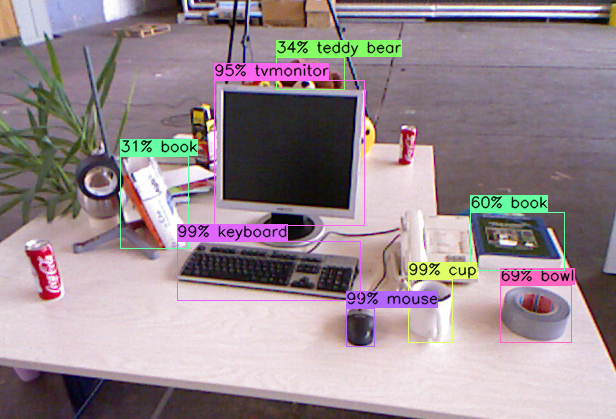}} 
		\label{fig:fig1a}
	}%
	\subfloat[\centering 3D map cloud with corresponding object spheres.]
	{\setlength{\fboxrule}{0.5pt}\setlength{\fboxsep}{0pt}\fbox{\includegraphics[width=0.47\linewidth]{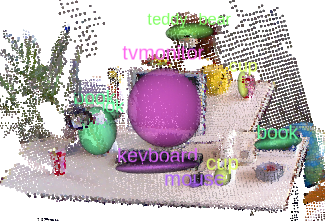}}		
		\label{fig:fig1b} 
	}%
	\caption{Example image of a point cloud map supplemented with 3D object markers that are derived from 2D detections. The markers are mapped as arbitrary colored spheres and bloated in their size by factor three for demonstration purposes.}
	\label{fig:fig1}
\end{figure}
\begin{figure*}[t]
	\centering
	
	\subfloat[Map from sequence \textit{fr2\_desk}]{%
		\begin{minipage}{0.32\linewidth}
			\includegraphics[width=1\linewidth]{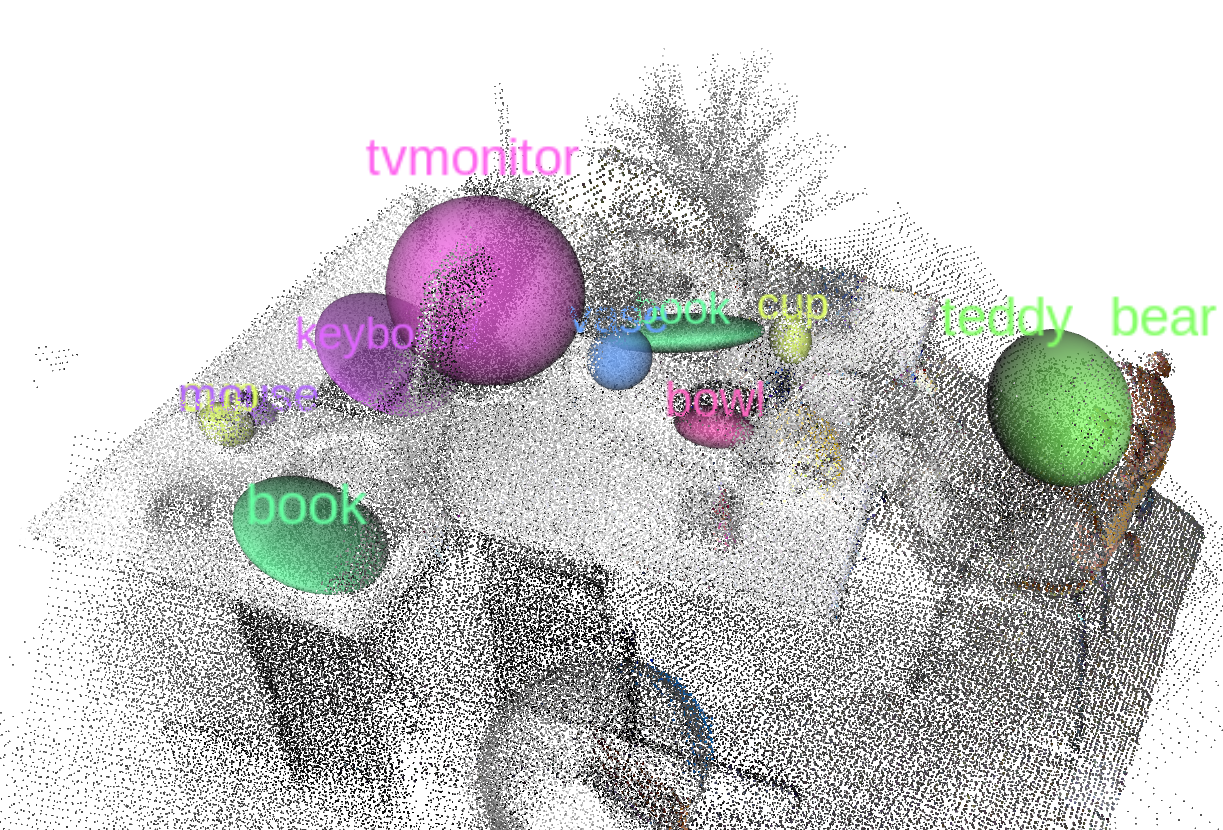}\\
		\end{minipage}
		\label{subfig:a}
		
	}
	\subfloat[Map from sequence \textit{fr3\_long\_office\_household}]{%
		\begin{minipage}{0.32\linewidth}
			\includegraphics[width=1\linewidth]{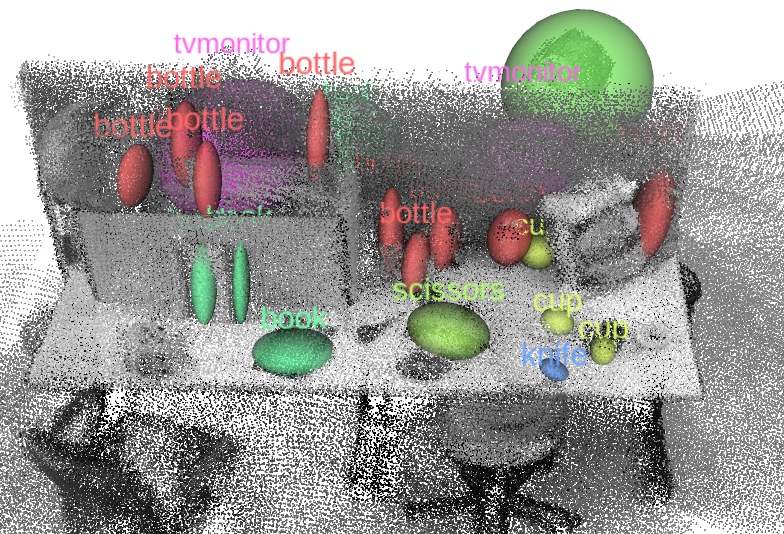}\\
		\end{minipage}
		\label{subfig:b}
	}
	\subfloat[Map from sequence \textit{fr3\_sitting\_xyz}]{%
		\begin{minipage}{0.32\linewidth}
			\includegraphics[width=1\linewidth]{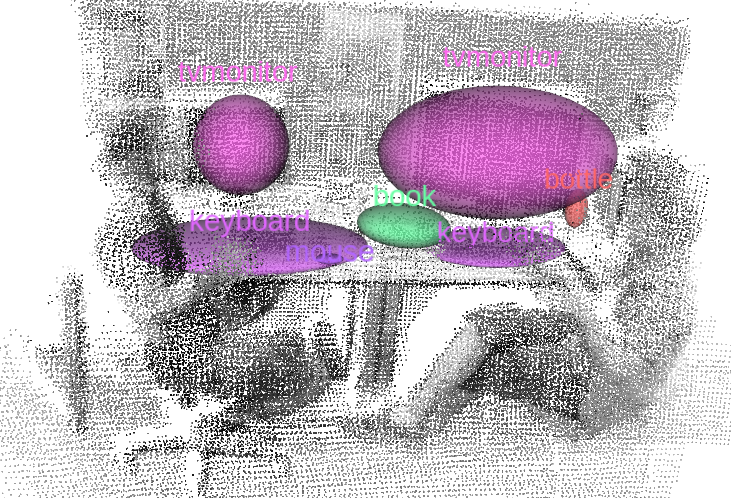}\\
		\end{minipage}
		\label{subfig:c}
	}
	\caption{Semantic mapping results from sequences of the RGB-D TUM dataset.}
	\label{fig:qualitative_results_tum}
\end{figure*}

\begin{table}[h!]
	\renewcommand{\arraystretch}{1.3}
	\caption{Applied system configuration for the experiments.}
	\label{table:Settings}
	\centering
	
	\begin{tabular}{l c }
		\hline
		Object Detections  \\ 
				\hline
		\textit{CNN confidence thresh} & 0.4 \\
		\textit{Min distance} & 0.2~m \\
		\textit{Max distance} & 25~m \\
		\hline 
		IOU Tracker  \\ 
				\hline
		\textit{$\sigma_{IOU}$ } & 0.2\\
		\textit{Min tracklet size} & 5\\
		\textit{Max time since last track det} & 0.5~s\\
		\hline
		Data Association \\
				\hline
		\textit{$u$} & 10\\ 		
		\hline
	\end{tabular}
\end{table}
\section{Experiments and Evaluation}
\label{experiments}

We evaluate our method's performance  qualitatively and quantitatively on the public available TUM RGB-D~(\citealt{rgbtum}) and stereo KITTI~(\citealt{Geiger2013IJRR}) datasets. 

\textbf{TUM RGB-D dataset}  The TUM RGB-D dataset consists of 39 real-world indoor sequences separated in categories such as \textit{Handheld SLAM}, \textit{Robot SLAM} or \textit{Dynamic Objects}. The sequences are recorded with an RGB-D Kinect camera in office environments with different levels of camera motion and varying environmental conditions such as dynamic objects and illumination changes. This gives us the excellent opportunity to examine our method against challenges such as image noise, motion blur, occlusion, and dynamic objects. 

\textbf{KITTI dataset}  The KITTI dataset provides 22 outdoor stereo sequences that were recorded from a ground vehicle in various large scale city, road and other urban environments. From these 22 sequences the first 10 are provided with ground truth odometry. Potential objects for semantic landmarks are mainly vehicles that are parking at the roadside. For proper mapping and localization our method has to deal with challenging occlusions and large object distances. 
 
We integrated the YOLOv4~(\citealt{yolo4}) network to provide us semantic object detections for every frame. It is trained on the COCO dataset~(\citealt{coco}) and able to differentiate between 80 different classes. A list of further system settings are shown in Table~\ref{table:Settings}. Their correct configuration depends on the scene and camera movement.Hectic camera trajectories, for example, demand loose IOU Tracker settings to ensure that detections can be associated to tracklets.  

Generated semantic landmarks are directly introduced into the GTSAM~(\citealt{gtsam}) graph optimizer backend to incorporate our detections into the pose optimization. We set a minimum distance of time between two landmark detection for the same object of \textit{2 seconds}.
All experiments were run a Linux system with AMD Ryzen 3950x CPU at 3.5~GHz and 64~GB of RAM and a NVIDIA RTX 2080 Ti.  

\subsection{Qualitative results}
\label{qualitative_results}
In the following, we will give some qualitative results for our proposed method based on sequences from the introduced datasets. 
\subsubsection{RGB-D TUM}
Figure~\ref{fig:fig1} shows a comparison between the input frame with predicted objects in Figure~\ref{fig:fig1a} and the corresponding point cloud map with enriched 3D objects in Figure~\ref{fig:fig1b}. Multiple items are successfully detected and projected as semantic spheres into the map. False positive predictions (\textit{e.g.} "bowl" for the tape) are rejected in our landmark proposal generation step. We were also able to adequately estimate the approximate objects size by their point clouds to determine their general dimensional appearances. Due to our landmark refinement step~(\ref{revaluation}) we are able to tweak our objects size estimation to converge it with the real size with the increasing information over the course of the sequence. An example of this is the teddy bear object sphere that is only encompassing its front face. The reason for this is that the landmark proposal that resulted in this object marker was generated from images at the beginning of the sequence where only a small part of the teddy bears face were visible (see Figure \ref{fig:fig1a}). As more landmark proposals are introduced from other points of view, the sphere size is increasing with every landmark fusion ending up completely covering the teddy bear. 
Figure \ref{fig:qualitative_results_tum} shows three more point clouds from TUM sequences that are extended with semantic markers from our generated landmarks, where Figure~\ref{subfig:c} was captured from the same sequence as Figure \ref{fig:fig1}. Here the teddy bears sphere has already increased in size as more candidate proposals had been fused into it.

Figure \ref{subfig:b} demonstrates that our method is also able to distinguish between small items of the same class represented by \textit{bottles} and \textit{cups} thanks to the adaptive validation gate. There is also a teddy bear present in this scenario (shown in the right-back). The capture was taken at the end of the sequence when the camera has already moved around the entire workstation. As a result, the teddy bear sphere entirely spans the object. 
 
\subsubsection{KITTI}
\begin{figure*}[t]
	\centering
	
	\subfloat[Map from sequence \textit{KITTI 00}]{%
		\begin{minipage}{0.49\linewidth}
			\includegraphics[width=1\linewidth]{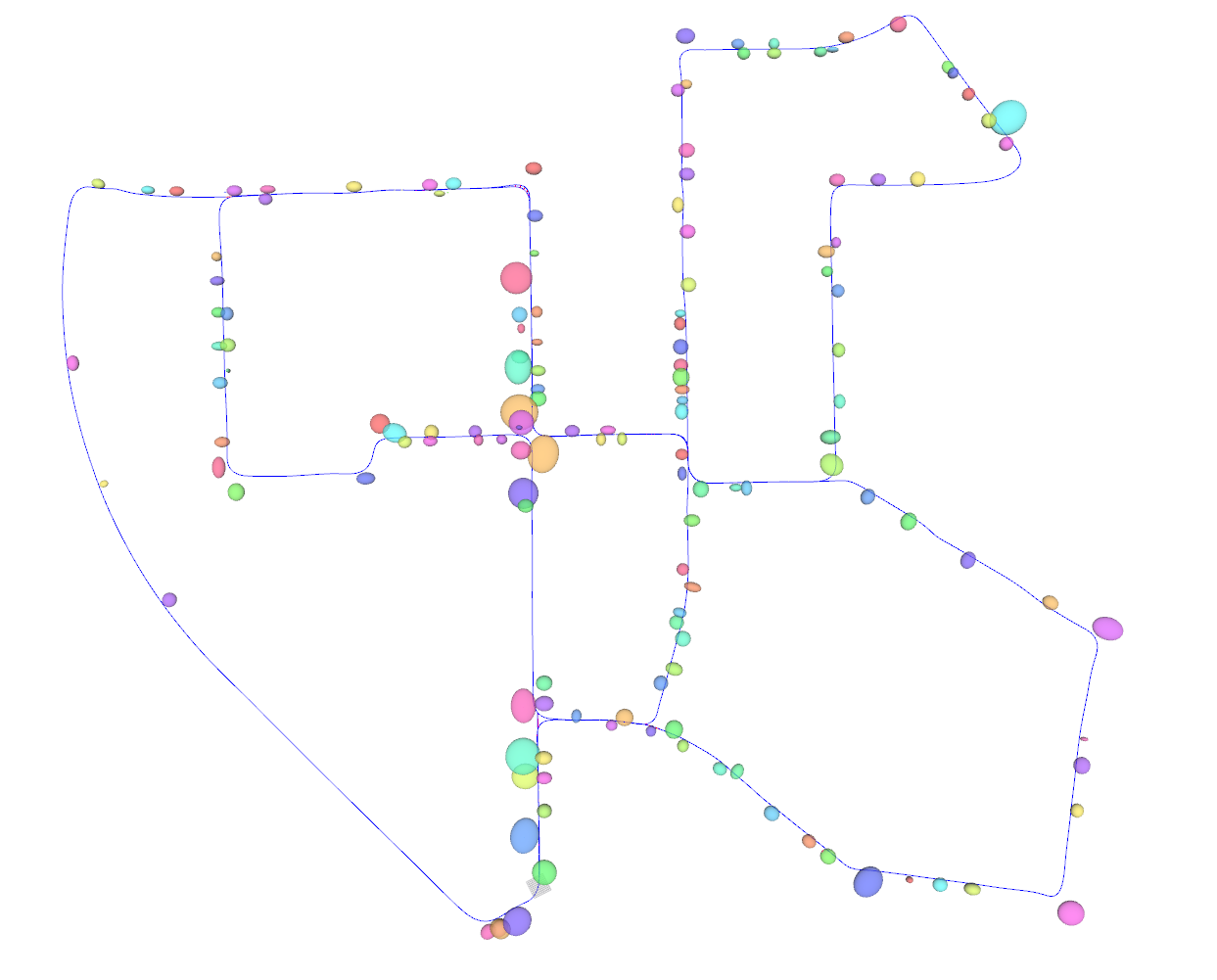}\\
		\end{minipage}
		\label{subfig:a_k}
		
	}
	\subfloat[Map from sequence \textit{KITTI 05}]{%
		\begin{minipage}{0.49\linewidth}
			\includegraphics[width=1\linewidth]{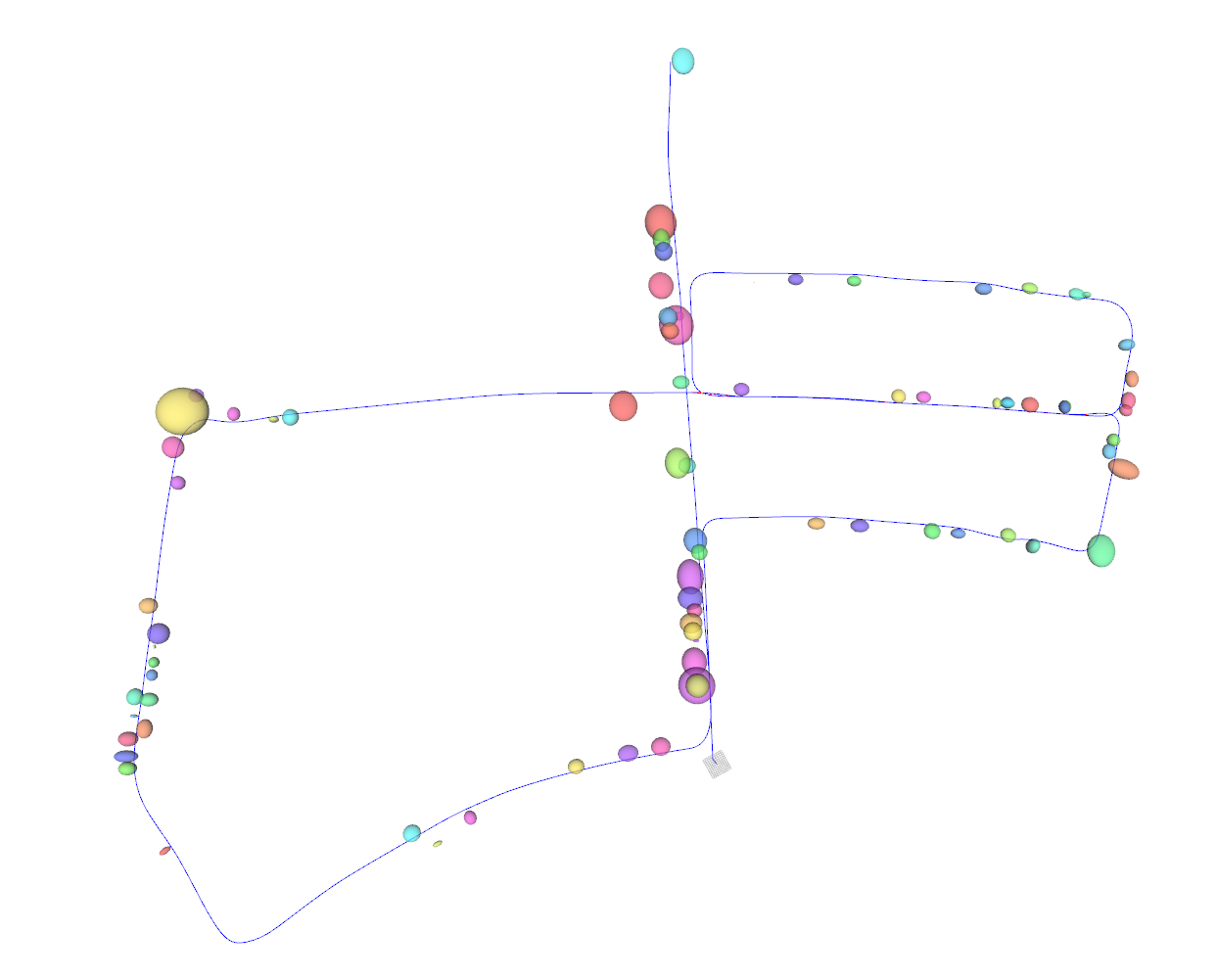}\\
		\end{minipage}
		\label{subfig:b_k}
	}
	\caption{Semantic mapping results of sequences from the KITTI dataset from top view.}
	\label{fig:qualitative_results_KITTI}
\end{figure*}

\begin{figure}[b!]
		\includegraphics[width=\linewidth]{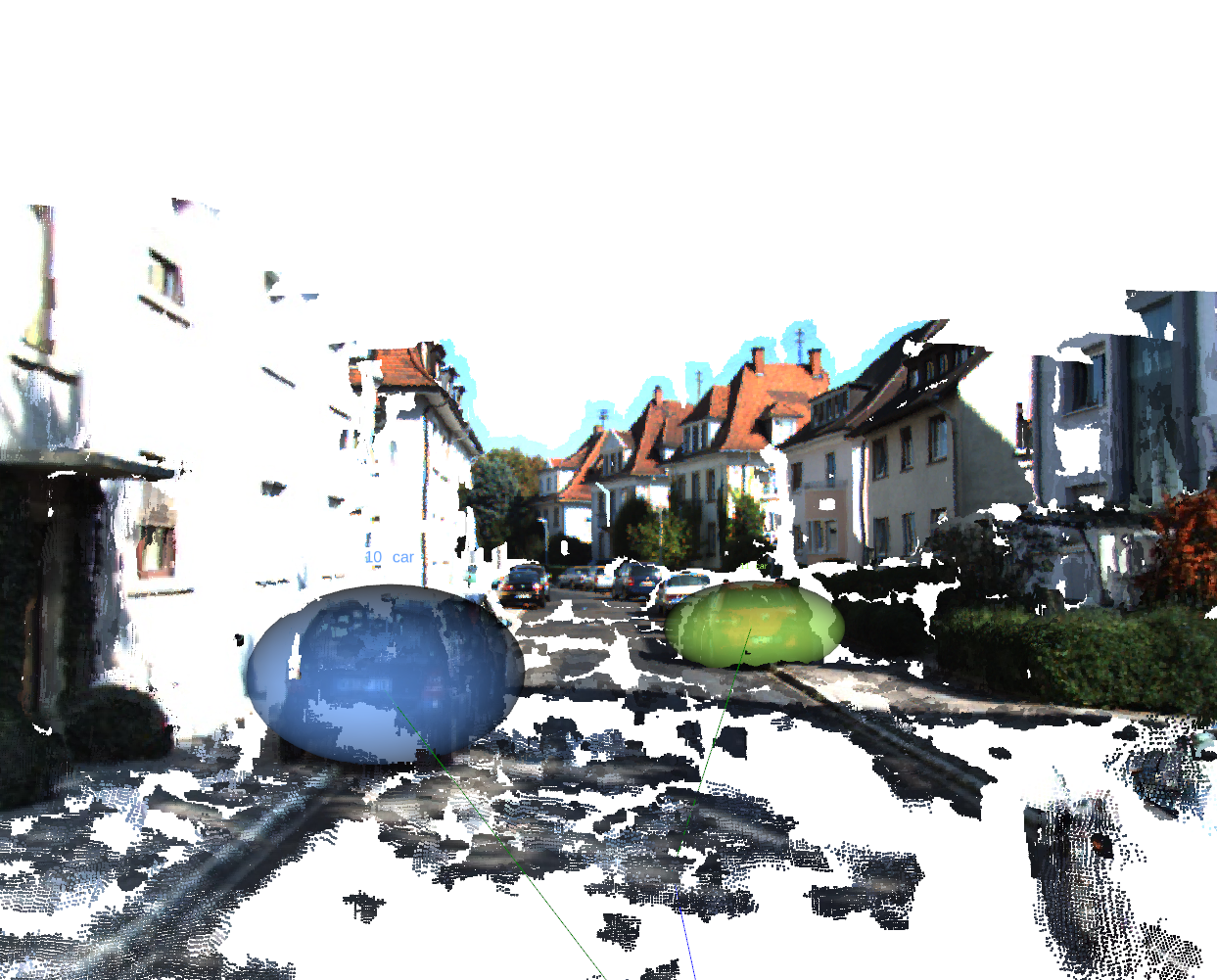}
		\caption{Generating object spheres with adapting size from sequence KITTI 00 from camera view.}
	\label{fig:quale_results_KITTI}
\end{figure}

In Figure \ref{fig:qualitative_results_KITTI} we showcase two exemplary trajectory maps from the KITTI dataset to demonstrate our method capability to perform semantic mapping in large scale. We removed the corresponding point cloud (in contrast to Figure \ref{fig:fig1_}) and increased the spheres by the factor 4 for improved visibility. The spheres are colored randomly. The first map from Figure \ref{subfig:a_k} is generated from the \textit{KITTI 00} sequence. In total 153 semantic objects have been successfully localized and added the map with the course of the scene. From these 153 objects, 150 are classified as \textit{cars}, 2 as \textit{trucks}, and 1 as \textit{motorbike}. Similarly, in Figure \ref{subfig:b_k} 80 semantic objects have been extracted from the \textit{KITTI 05} sequence, that are divided into 72 cars and 8 trucks.
Figure \ref{fig:quale_results_KITTI} shows an example image capture from the camera point of view in the middle of the sequence, where two cars getting added as semantic landmarks. The sphere's size demonstrates that our method is able to precisely estimate the positions and sizes of the objects.

\subsection{Quantitative results}
We perform multiple experiments with variable settings and scenarios to quantitatively evaluate or methods performance of deriving useful semantic constraints for the localization. We chose the commonly used absolute trajectory root mean square error (RMSE) as error metric. The absolute trajectory error between the estimated trajectory and the corresponding ground truth trajectory at time $i$ is defined as
\begin{equation}
E_i:=Q_i^{-1}SP_i.
\end{equation}
$S$ is a rigid-body transformation that maps the estimated trajectory $P$ onto the ground truth trajectory $Q$. The error of the entire sequence is the average deviation from the ground truth trajectory per frame:
\begin{equation}
ATE_{rmse}:=\left( \frac{1}{n} \sum_{i=1}^{n}||trans(E_i)||^2 \right) ^{\frac{1}{2}}
\end{equation}

\subsubsection{RGB-D TUM}
\label{quant_rgbd}
We selected in total 9 sequences from categories \textit{Handheld SLAM} and \textit{Dynamic Objects} that are commonly used for SLAM related approaches. Three of them (\textit{fr1\_desk, fr3\_long\_office\_household, fr2\_desk}) show office environments recorded from a handheld camera with multiple objects placed on the desks. In the other sequences the camera films a desk with people sitting or walking around it. The camera performs different types of rotational and translational motions, while the people serve as dynamic objects. The combination of alternating ego-motion and a dynamic environment makes these sequences very challenging for evaluation the robustness and precision of the trajectory estimation. Table \ref{table:comp_orb_slam} shows the localization errors for every experiment with ORB-SLAM2 solely and ORB-SLAM2 in combination with our semantic mapping (\textit{SM}) method.
We use the median of five iterations for each configuration to reduce the non-deterministic impact from the odometry estimator on the results, introduced by the ORB feature extraction. It shows that our method is able to improve the trajectory estimation in scenes where a vigorous dynamic environment strongly decreases the general performance of the baseline. While the base localization uses all extracted features including those laying on the moving persons in the scene, our methods tracklet approach prevents the creation of semantic landmarks from these moving objects. As a result, the landmark detections for the pose graph optimization were only created from the static objects on the desk in the background leading to a correction of the corrupted pose estimation from the ORB-SLAM2. Especially in the \textit{walking\_xyz} sequence our approach demonstrates that the tracking of the items on the desk and utilizing them for pose optimization helps to correct the trajectory drift caused by the dynamic objects.

\begin{table}[t]
	\renewcommand{\arraystretch}{1.3}
	\caption{RMSE of the absolute trajectory error [m] for TUM sequences.}
	\label{table:comp_orb_slam}
	\centering
	
	\begin{tabular}{l c c c}
		\hline
		Sequence & ORB-SLAM2 & ORB-SLAM2 & Impr.  \\
		& & + SM \\ \hline 
		\textit{fr1\_desk} & 0.052 & \textbf{0.042} & \textbf{19.2}~\% \\
		\textit{fr2\_desk} & 0.072 &  \textbf{0.047} & \textbf{34.7}~\% \\
		\hline
		\textit{fr2dwp} & 0.081 & \textbf{0.062} & \textbf{23.5}~\%  \\
		\textit{fr3s\_xyz} & \textbf{0.012} & 0.014 &  -16.7~\% \\
		\textit{fr3w\_static} & \textbf{0.042} & 0.047 & -11.9~\%\\
		\textit{fr3w\_hs} & 0.259 & \textbf{0.171} & \textbf{34.0}~\%   \\
		\textit{fr3w\_rpy} & 0.457 & \textbf{0.377} & \textbf{17.5}~\% \\
		\textit{fr3w\_xyz} & 0.387 & \textbf{0.124} & \textbf{68.0}~\%  \\		
		\hline
	\end{tabular}
\end{table}

\begin{table}[b!]
	\renewcommand{\arraystretch}{1.3}
	\caption{RMSE of the absolute trajectory error [m] for KITTI sequences.}
	\label{table:KITTI}
	\centering
	
	\begin{tabular}{l c c c c}
		\hline
		Sequence & 00 & 05 &  06 & 07 \\ \hline 
		\textit{\cite{Doherty2020ProbabilisticDA}} & - & 5.718 &-  & -\\
		\textit{\cite{slam:bernreiter}} & 4.54 & 4.4 & \textbf{2.3} & 2.9 \\
		\hline
		\textit{ORB-SLAM2} & 4.841 &  3.263 & 3.554 &  1.929 \\
		\textit{ORB-SLAM2 + LC  }& \textbf{4.228} & 3.245 & 3.428& 1.801\\
		\textit{ORB-SLAM2 + SM} & 4.495 & 3.202 & 3.438& 1.808\\ 
		\textit{ORB-SLAM2 + SM + LC}& 4.389 & \textbf{3.162} & 3.043 & \textbf{1.775}\\
		
		\hline
	\end{tabular}
\end{table}
In scenes with a static and feature-rich environment the ATE could not be improved by our method in a comprehensive manner. One reason is that the trajectory estimation of the baseline is already very accurate. Another reason is that we tend to provide landmark data with small inaccuracies for the graph optimization, when we use the centroid of the detected object as landmark location. Unfortunately, our calculated centroid is only an estimation of the real centroid as we can only capture an extract of the whole object as already mentioned in section~\ref{qualitative_results}. Hereby, we introduce small errors into the graph optimization. This error size varies with the size of the object and therefore the difference between the estimated and the real centroid of the landmark. That makes big objects like tables less useful for incorporating into the localization process.  
In sequences with general low ATE even small errors from the centroid estimation can result into a performance loss as shown for \textit{fr3\_long\_office\_household}. 

\subsubsection{KITTI}
We selected 4 scenes from the KITTI dataset. In this scenes the trajectory performs loop closures which gives us the opportunity the examine if our method is able to identify prior mapped objects when the trajectory returns to a know place and if resulting semantic constraints improve the trajectory accuracy. We compare our result against two other semantic sensitive SLAM frameworks from \cite{Doherty2020ProbabilisticDA} and \cite{slam:bernreiter}. Additionally, our tests are executed using different combinations with loop closure (\textit{LC}) detection to examine our semantic mapping performance and interplay with it. 

In comparison the stereo ORB-SLAM2 without loop closure and semantic constraints provides the worst performance as the drift accumulates over time and doesn't get corrected in the following path. However, running the sequence with loop closure or semantic mapping leads to an improvement, whereas loop closure algorithm performs in most sequences slightly better than the semantic mapping. The reason is that the loop closures in every sequence occur from the same point of view and, thus, provide ideal preconditions for 2D feature-based loop closure detection methods. Using, both, loop closure detection and semantic mapping results in the smallest ATE except in one case, whereas in comparison with the other approaches \cite{slam:bernreiter} provides slightly better results in sequence 6. In the first sequence the combination of loop closure detection and semantic mapping performs somewhat worse than loop closure detection solely. We found that, similar to the discussion in \ref{quant_rgbd}, this comes from a semantic detection that was associated with a prior landmark, where the centroid was estimated from a different point of view. As a result, the small distance between both estimated centroids lead to an insufficient trajectory optimization.

\subsection{Runtime analysis}
For the useful application along with other real-time SLAM implementations our method has to be equally computational efficient. Table~\ref{table:runtime} shows the average processing time for each step measured over the \textit{fr2\_desk} sequence. An average cycle of our main thread takes around 65 ms. In some occasions, where the association using the validation doesn't lead to a clear decision, the second association layer (\textit{Candidate association 2}) with its cloud based nearest neighbor method is performed. In this case, roughly 150~ms are additionally needed for the association decision. Object detections together with the candidate proposal step are processed in a separate thread. This makes sure it can efficiently find valid object proposals over consecutive frames without being slowed down by the time consuming association step. Generally, the processing-time is strongly affected by multiple factors. Small objects represented by a smaller point cloud are faster handled than bigger objects. The more potential association candidates are within the validation gate of new a proposal the longer the decision takes. In our experiments, our system ran seamlessly besides RTAB-Map and was able to provide multiple detections for every keyframe estimation.

\begin{table}[t]
	\renewcommand{\arraystretch}{1.3}
	\caption{Average runtimes for each process stage over one entire iteration recorded on the \textit{fr2\_desk} sequence. }
	\label{table:runtime}
	\centering
	
	\begin{threeparttable}
	\begin{tabular}{l c }
		\hline
		Process & Time (ms)  \\ \hline 
		\textit{Object detection} & 22.62 \\
		\textit{Candidate proposal} & 0.069\\	
		\textit{Candidate association 1 / + 2\tnote{*}} & 7.8 / + 150\\
		\textit{Landmark update \& merge} & 23.16 \\
		\hline
	\end{tabular}
  \begin{tablenotes}
	\item[*] Only executed if association method 1 doesn't provide sufficient results.
\end{tablenotes}
\end{threeparttable}
\end{table}

\subsection{Limitation}
We perform a deterministic and greedy data association, which enables real-time application capabilities and a wide variety of possible applications (\textit{e.g.} using AprilTag interface). These benefits are accompanied by the disadvantage that possible errors in the data association can have a strong negative impact on the validity of the odometry estimation.   
Another difficulty are objects of the same class that are placed very close to each other. In this case, the validation gate has to be smaller than the distance between in order to avoid getting treated as one single object. 
A possible future solution to tackle these limitations could be the implication of probabilistic approaches for accepting only highly confident associations and holding back those candidates until their assignment is approved by subsequent detections.

\section{Conclusions}
\label{conclusion}

We presented a method for extending SLAM methods with semantic objects in a robust and efficient manner. While our generated semantics increase the perceptual understanding of the environment, we additional provide an approach for object association to use them as static landmarks for trajectory optimizations. This enables the SLAM system to be used in a wider range and more complex applications such as Human-Robot Interaction and object manipulation for mobile platforms. At the same time, by rejecting dynamic objects as semantics our method is able to rectify the trajectory from the defective influence of dynamic objects and other disruptive conditions on the odometry estimation so that these applications can be performed in more challenging environments.

To ensure that our approach has the capabilities to work along with other components in real applicable scenarios, we designed our system to operate in real-time.
As further research, we will tackle shortcomings in the challenging data association by introducing a probabilistic approach where multiple hypotheses will be considered before taking an association decision. 

Another direction is to extend our approach with additional processing steps for estimating object poses. This will be equally useful for the data association as well as for real application scenarios for mobile robots.

\section*{Acknowledgement}
This work was funded by the German Federal Ministry of Education and Research (BMBF)
under Grant Nos. 03ZZ0448L (RoboAssist) and 03ZZ04X02B (RoboLab) within
the Zwanzig20 Alliance 3Dsensation. 

\bibliography{bibliography.bib}

\end{document}